\pdfoutput=1
\documentclass[11pt]{article}
\usepackage{emnlp2021}
\usepackage{times}
\usepackage{latexsym}
\usepackage{graphicx}
\graphicspath{ {./Figs/} }
\usepackage[T1]{fontenc}
\usepackage[utf8]{inputenc}
\usepackage{microtype}
\usepackage{pifont}
\usepackage{latexsym}

\usepackage{balance}
\usepackage{helvet}  
\usepackage{courier}  
\usepackage{url}  
\usepackage{graphicx}  
\usepackage{amssymb}
\usepackage{amsmath}
\usepackage{algorithm}
\usepackage{algorithmic}
\usepackage{amsthm}
\usepackage{multirow}
\usepackage{pgffor}
\usepackage{graphicx}
\usepackage{epstopdf}
\usepackage{tikz}
\usepackage{epstopdf}
\usepackage{mathtools}
\usepackage{enumitem}
\usepackage{subfigure}
\usepackage{tabulary}
\usepackage{color}
\usepackage{url}
\usepackage{flushend}
\usepackage{stfloats}
\usepackage{tcolorbox}
\usepackage[utf8]{inputenc}
\usepackage[english]{babel}
\usepackage{tabularx}
\usepackage{CJKutf8}
\usepackage{microtype}
\usepackage{textcomp}
\usepackage{booktabs}
\usepackage{colortbl}
\usepackage{soul}
\usepackage{arydshln}
\usepackage{natbib}
\usepackage{appendix}
\def\aclpaperid{3684}
\usepackage{xspace}
\usepackage{microtype}
\usepackage{balance}
\usepackage{flushend}
 
\newcommand{\ourapproach}{\textsc{AMG}\xspace}

\title{Attend, Memorize and Generate: Towards Faithful Table-to-Text Generation in Few Shots}

\author{
  \textbf{Wenting Zhao}$^1$~~~~ \textbf{Ye Liu}$^1$~~~~\textbf{Yao Wan}$^2$~~~~  \textbf{Philip S. Yu}$^1$\\
$^1$ Department of Computer Science, University of Illinois at Chicago, IL, USA \\
  $^2$School of Computer Sci. \& Tech., Huazhong University of Science and Technology, China\\
  { \texttt{\{wzhao41, yliu279, psyu\}@uic.edu}, \texttt{wanyao@hust.edu.cn}} \\
}
\begin{document}
\maketitle
\begin{abstract}
Few-shot table-to-text generation is a task of composing fluent and faithful sentences to convey table content using limited data.
Despite many efforts having been made towards generating impressive fluent sentences by fine-tuning powerful pre-trained language models, the faithfulness of generated content still needs to be improved.
To this end, this paper proposes a novel approach 
\textit{\underline{A}ttend, \underline{M}emorize and \underline{G}enerate} (called \ourapproach), inspired by the text generation process of humans.
In particular, \ourapproach (1) attends over the multi-granularity of context using a novel strategy based on table slot level and traditional token-by-token level attention to exploit both the table structure and natural linguistic information; (2) dynamically memorizes the table slot allocation states; and (3) generates faithful sentences according to both the context and memory allocation states. 
Comprehensive experiments with human evaluation on three domains (i.e., humans, songs, and books) of the \textit{Wiki} dataset show that our model can generate higher qualified texts when compared with several state-of-the-art baselines, in both fluency and faithfulness.\footnote{All the source code and experimental dataset are available at \url{https://github.com/wentinghome/AMG}.}

\end{abstract}

\section{Introduction}

Table-to-text generation, which aims to translate a semi-structured table into natural language descriptions while preserving the conveyed table information, 
are drawing increasing interest over the past few years.
It has been widely applied in many real-world scenarios, such as automatically generating weather forecasting reports~\citep{liang-etal-2009-learning}, biographies~\citep{lebret2016neural,wang-etal-2018-describing}, restaurant descriptions~\citep{novikova-etal-2017-e2e}, task-oriented conversations~\citep{budzianowski2018multiwoz,williams-etal-2013-dialog} as well as healthcare descriptions~\citep{dimarco2007development,hasan2019clinical}. 
Despite such significant gains, current approaches are driven by large-scale well-labeled training data, 
hindering the generalization to other scenarios with limited labeled data. In addition, the faithfulness of generated contents is still not well explored.
\begin{figure}[t]
    \centering
    \includegraphics[width=0.48\textwidth]{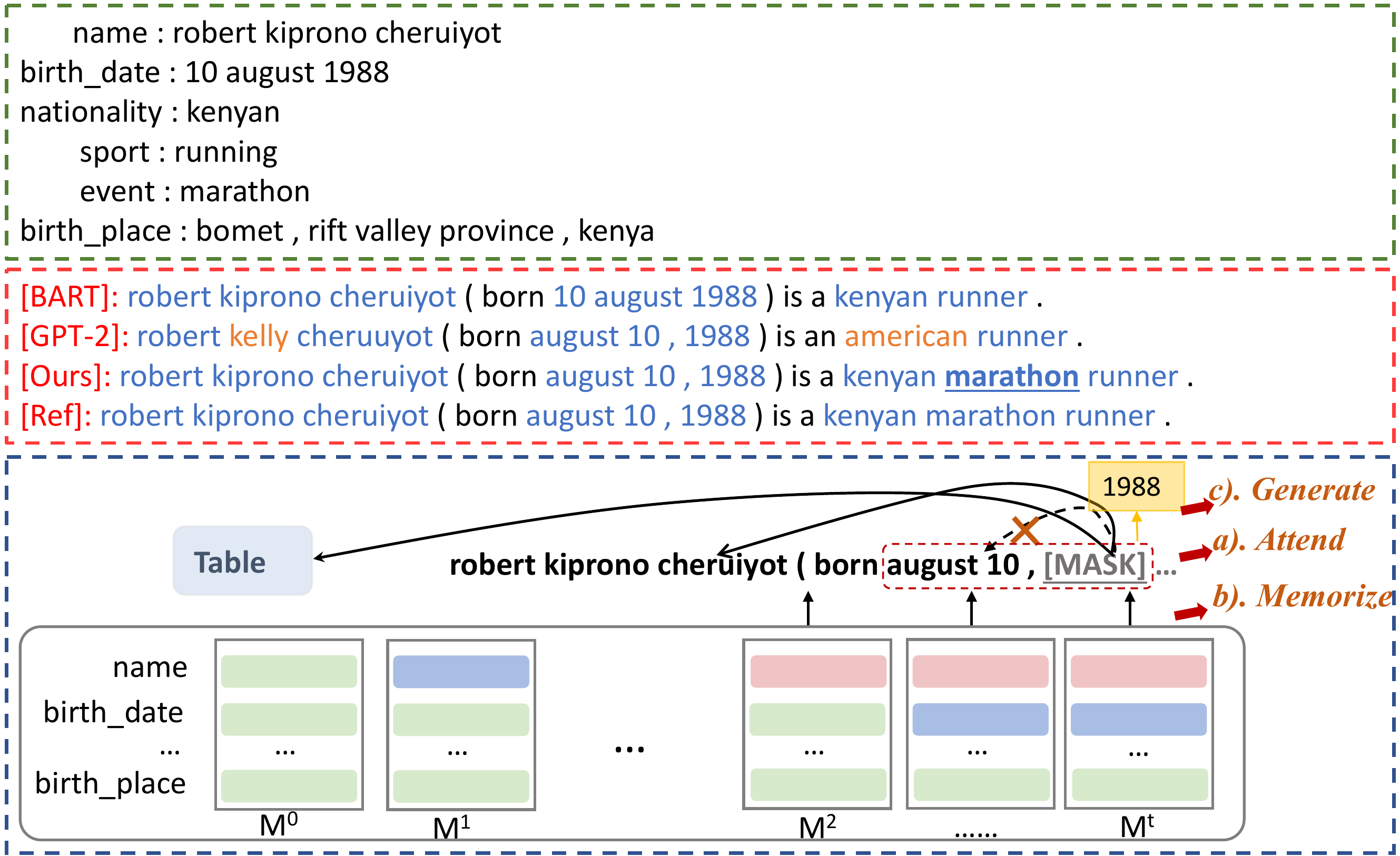}
    \caption{A motivating example.}
    \label{fig:intro_1}
\end{figure}

Few-shot natural language generation~\citep{brown2020language, schick-schutze-2021-just, xia-etal-2020-composed} has been in increasing demand since sufficient labeled data are always unavailable in many scenarios.
To improve the table-to-text generation in few-shot scenarios, 
many existing works \citep{chen-etal-2020-shot,gong-etal-2020-tablegpt,peng-etal-2020-shot} resort to the pre-training techniques which have been widely adopted in NLP, that is, pre-training a model first on large-scale unlabeled data, and then transfer the learned knowledge in pre-trained model to the few-shot scenario of table-to-text generation. 
Although these pre-trained models have achieved promising performance on generating fluent descriptions, from our investigation, they are still suffering from three major limitations: 
(1) \textit{The structure of table has not been well preserved.}
On table representation, existing methods~\citep{chen-etal-2020-shot,gong-etal-2020-tablegpt,chen-etal-2020-logical} used to flatten the table into sequential sentences, ignoring the structured features (e.g., correlation between words within each table slot) among tables, which is also critical for table-to-text generation. 
(2) \textit{Generation bias.} Current approaches that directly fine-tune the model on target data make the model in favor of the knowledge learned from pre-training rather than specific target task knowledge, hurting the faithfulness because extra information irrelevant to the input table is introduced.

For example, as shown in Figure~\ref{fig:intro_1}, given a table in the top box, the aim is to generate a coherent and faithful sentence with high coverage of table slots,  as well as less out-of-table information. From this table, we can observe that current state-of-the-art models tend to generate sentences with hallucinated contents. For example, GPT-2 introduces wrong middle name ``\textit{kelly}'' and the nationality ``\textit{american}''. In addition, the table coverage of contents generated by current approaches is low. For example, BART does not mention the event ``\textit{marathon}''.
These observation motivate us to design a model that can generate faithful texts from tables while keeping the fluency.

To tackle the aforementioned limitations, this paper proposes a novel approach \textit{\underline{A}ttend, \underline{M}emorize and \underline{G}enerate} (called AMG) for faithful table-to-text generation in few-shots.
Inspired by the human generation process which copies a consecutive slot span to compose a sentence using the context, we propose a table slot attention mechanism to empower the model generalization ability in inference by strengthening the dependency between the generated sentence with the input table.
In addition, to avoid generating hallucinated contents, we design a memory unit to monitor the visits of each table slot.
Particularly, the memory unit is initialized as all the meta-data of table slots, and then updated by checking the generated words as well as the current memory state. 

Looking back to Figure~\ref{fig:intro_1}, we can also observe several advantages of \ourapproach.
First of all, we can see \ourapproach allows the to-be-predicted word ``\textit{1998}'' from ``\textit{birth\_date}'' table slot to attend on the table as well as the previously generated sentence ``\textit{robert \dots born}'', while the attention on within table slot words are prohibited. Thus, the model is enforced to capture the table span structure and rely on the table span value to generate. To this end, the model learns to capture the slot level table representation.

Furthermore, as shown in Figure~\ref{fig:intro_1}, ``\textit{$M^0$}'' is the memory initial state where all the slot are available to be chosen (marked by green). After predicting the last word of table slot ``\textit{name}'', ``\textit{$M^1$}'' will be updated since it detects that the table slot ``\textit{name}'' is present in the generated sentence, thus making the state of ``\textit{name}'' unavailable (marked by red). In addition, the generation of word ``\textit{1998}'' takes the context and table slot allocation into account, therefore ``\textit{1998}'' is selected by locating the value of table span ``\textit{birth\_date}'' as well as the activated signal of table slot ``\textit{birth\_date}'' (marked by blue) from memory allocation status.

To summarize, the primary contributions of this paper are as follows:
    (1) To better preserve the structure of table, we design a multi-grain attention that can attend over the table word as well as table slots level.
    (2) It is the first time that we introduce a memory mechanism to improve the faithfulness of generated texts by tracking the allocation of table slots.
    (3) We have conducted comprehensive experiments on three domains (i.e., Humans, Books and Songs) of the \textit{Wiki} dataset to validate the effectiveness of our proposed approach.

\section{Preliminaries}
\subsection{Problem Definition}
Given a table $T$ of $m$ attribute-value pairs $\{(a_i, v_i)\}_{i=1}^{m}$, where $a_i$ and $v_i$ refer to the attribute name and value of $i$-th table slot, respectively, the table-to-text generation task aims at producing a coherent text $Y=(y_1,\cdots,y_L)$ that can describe the table information with fluency and faithfulness, where $L$ denotes the length of generated text. 

\subsection{UniLM}
To alleviate the under-fitting issue caused by insufficient training examples in few shot learning, \ourapproach adopts the state-of-art pre-trained language model UniLM~\citep{dong2019unified} structure to integrate the external knowledge.
UniLM is a multi-layer Transformer network which can be applied into both tasks of natural language understanding (NLU) and natural language generation (NLG). In this paper, we configure UniLM using Seq2Seq self-attention mask to aggregate the context of the masked $i$-th to-be-predicted word $y_{i}^{[MASK]}$ that are source sequence words from table $T$, and the previously generated target words $y_{<i}$. The proposed model computes the conditional probability for the to-be-predicted word using the masked language model objective function, as follows:
\begin{equation}
\label{eq:mem-update-1}
P(Y|T; \theta) = \prod_{i=1}^{L}P(y_{i}^{[MASK]}|y_{<i},T ; \theta)\,.
\end{equation}

\section{\ourapproach Approach}

\subsection{Overview}

\begin{figure*}[t]
\centering
\includegraphics[width=0.86\textwidth]{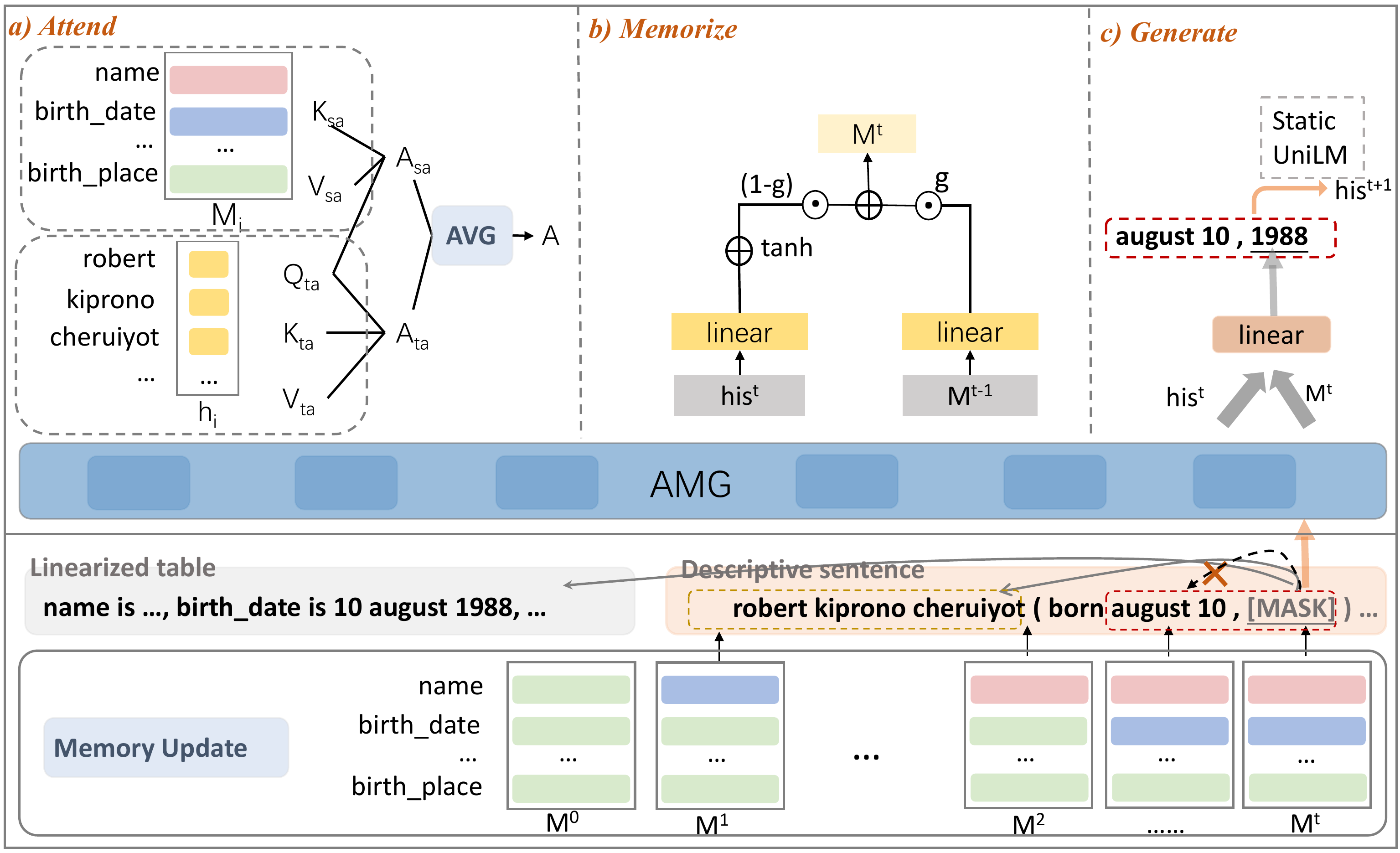}
\caption{An overview of \ourapproach. The input to \ourapproach is the concatenation of linearized table (marked in grey) and the descriptive sentence(marked in orange). The bottom box shows the memory update process. The top three boxes show the building blocks of \ourapproach, designed to attend, memorize and generate descriptions from tables.}
\label{fig:model_overview}
\centering
\end{figure*}

Figure~\ref{fig:model_overview} illustrates the overall architecture of our model, which is composed of three components, i.e., \textit{attend}, \textit{memorize}, and \textit{generate}.
(1) \textit{Attend.} We propose a multi-granularity attention mechanism which attends over both token level and the table slot level to capture the linguistic knowledge as well as table structure information.
We think that these knowledge can improve the faithfulness of generated texts.
(2) \textit{Memory.} We develop a memory to store and keep track of the table slot allocation status.
(3) \textit{Generate.} We take both the context representation and the table slot allocation states into account while making predictions. 
The above three building blocks interweave and lead the model to generate descriptions from tables faithfully.


\subsection{Table Representation}
\paragraph{Table Linearization}
Table-to-text generation receives semi-structured table as input. However, our proposed model \ourapproach is built upon the UniLM architecture which requires natural sentence as input. Therefore, the first step we need to do is to translate the table into a natural sentence by linearization~\cite{chen-etal-2020-shot}. For the table example shown in Figure~\ref{fig:intro_1}, the attribute value pair ``\textit{name: robert kiprono cheruiyot}'' can be linearized as ``\textit{name is \texttt{[E\_CLS]} robert kiprono cheruiyot \texttt{[E\_SEP]};}'', where \texttt{[E\_CLS]} and \texttt{[E\_SEP]} are two special tokens to indicate the beginning and the end of table slot value. 

\paragraph{Representing the History of Table Slot Allocation}

\ourapproach makes prediction on the to-be-predicted token by taking the memory allocation status into account. The memory at different time step is updated by the previously generated table slots. Thus, we need to prepare the previously generated table slot representation $his^{t}$ at time step $t$ by using the static UniLM model. For example, in Figure~\ref{fig:model_overview}, when making prediction for ``\texttt{[MASK]}'', the representation of table slot allocation history is computed by feeding ``\textit{robert kiprono cheruiyot}'' to the static UniLM model and obtain the average of hidden states.

\subsection{Multi-Granularity Attention}
\ourapproach introduces the multi-granularity attention (MA) which is the combination of two granularity of attention, i.e., token level and table slot level attention. 
The token level attention is the original UniLM token level attention while the table slot level attention is the extra attention over table slot memory.
The advantage is that the table slot attention can provide an extra signal to the UniLM, encouraging \ourapproach to copy tokens from the table slot value that have not appeared in the target. 
As shown in Figure~\ref{fig:model_overview}, the memory augmented attention $A$ is the average of token level attention $A_{ta}$ and table slot level attention $A_{sa}$, as following:
\begin{equation} 
\label{eq:mem_augmented_att}
A = (A_{ta} + A_{sa})/2\,,
\end{equation}
where the token level self-attention mechanism learns a unique series of query matrix $W_{Q_{ta}}^{l}$, key matrix $W_{K_{ta}}^{l}$, and value matrix $W_{V_{ta}}^{l}$ at the $l$-th Transformer layer for each attention head. Then, \ourapproach maps the $(l-1)$-th Transformer layer output $T^{l-1}$ to three matrices: query $Q_{ta}$, key $K_{ta}$, and value $V_{ta}$. 
The output of a self-attention head $A_{ta}$ is computed as Eq.(\ref{eq:self-att}), where $Mask^{ta}\in \mathbb{R}^{N \times N}$ is the seq2seq attention mask, allowing the to-be-predicted token to attend to table tokens as well as the previously generated tokens. $N$ refers to the total token length of table, previously generated tokens and the current to-be-predicted token.
\begin{equation} 
\label{eq:self-att}
A_{ta}^{l} = \operatorname{softmax}(\frac{Q_{ta}K_{ta}^T}{\sqrt{d^k}}+Mask^{ta})\cdot V_{ta}\,, 
\end{equation}

\paragraph{Table Slot Attention}

Table slot attention works in a similar way with the self attention, while the major difference is to learn new key and value mapping matrices $W_{K_{sa}}^{l}$ and $W_{V_{sa}}^{l}$ and project memory $M^{l-1}$ using $W_{K_{sa}}^{l}$ and $W_{V_{sa}}^{l}$ to obtain $K_{sa}$ and $V_{sa}$. The query $Q_{sa}$ is computed by the projection of UniLM hidden state $h^{l-1}$ using mapping matrix $W_{Q_{sa}}^{l}$.
Memory $M$ in \ourapproach is defined as a $\mathbb{R}^{d_h \times slot_n}$ matrix where $slot_n$ is the maximum number of table slots. The $j$-th column of memory at time step $t$ is denoted as $M_{j}^{t}$, and the initial state of memory $M_{j}^{0}$ is the average embedding of the $j$-th table slot value computed using static UniLM model. The output of slot level attention head $A_{sa}^{l}$ is as follows:
\begin{equation} \label{eq:mem-attention}
    \begin{split}
    & Q_{sa} = h^{l-1}W_{Q_{sa}}^{l} \\
    & K_{sa} = M^{l-1}W_{K_{sa}}^{l} \\
    & V_{sa} = M^{l-1}W_{V_{sa}}^{l} \\
    & A_{sa}^{l} = \operatorname{softmax}(\frac{Q_{sa}K_{sa}^T}{\sqrt{d^k}}+Mask^{slot})\cdot V_{sa}\,.
    \end{split}
\end{equation}
Instead of applying the original seq2seq attention from UniLM to the input, a table slot attention mask $Mask^{slot}\in \mathbb{R}^{N \times N}$ is introduced to decide which word should be attended. In our case, we prohibit the to-be-predicted token to attend the previously generated words within the same table slots, while allow to attend the rest of generated words and the table. As shown in Figure~\ref{fig:model_overview}, ``\textit{1998}'' from the descriptive sentence can attend to both the table ``\textit{
name is \dots, birth\_date is \dots}'' and previously generated words ``\textit{robert kiprono cheruiyot ( born}'', while is not allowed to attend to words within the same table slot ``\textit{august 10 ,}''.

\paragraph{Table Slot Memory Update}
\ourapproach updates the memory matrix multiple times dynamically depending on how many times the generated sentence finishes generating one entire table slot value. To give a clear signal for the model to detect the beginning and the end of the table slot value, we introduce two additional special tokens \texttt{[E\_CLS]} and \texttt{[E\_SEP]} into the reference. Memory is updated using the gated mechanism, following~\cite{henaff2016tracking}:
\begin{equation} \label{eq:mem-update}
    \begin{split}
    & \hat{M}_j^t = \operatorname{tanh}(W_aM_j^{t-1} + W_{b}his^{t-1}) \\
    & z_j^t = \delta(W_cM_j^{t-1} + W_{d}his^{t-1}) \\
    & M_j^{t} = (1-z_j^t)  M_j^{t-1} + z_j^t  \hat{M}_j^t\,.
    \end{split}
\end{equation}
In Eq.(\ref{eq:mem-update}), $W_a$, $W_b$, $W_c$ and $W_d$ are trainable parameters. First, $\hat{M}_j^t $ is the new candidate memory to be combined with the existing memory $M_j^{t-1}$. Then, the gate function $z_j^t$ employs a sigmoid function $\delta$ to determine how much memory $M_j^{t}$ will be influenced. At last, we retain $M_j^{t}$ by using gate function to control how much each cell in memory is updated by considering the history of table slot appearance in the target sentence, as well as the last memory.

\paragraph{Text Generation} When predicting the next token at each time step, \ourapproach considers both the context representation and the table slot allocation status from memory shown in Eq.(\ref{eq:generate}) where $tb$ refers to the table representation, $tk^t$ denotes the token predicted at time $t$ by \ourapproach, and $tk^{0 \dots t-1}$ denote the tokens previously generated from time $0$ to $t-1$.
\begin{multline} \label{eq:generate}
    (his^{t},M^{t},{tk}^{t}) =\\ \ourapproach(tb, his^{t-1},M^{t-1}, tk^{0 \dots t-1})\,.  
\end{multline}

\subsection{Task-Adaptive Pre-Training}
\ourapproach is built upon the pre-trained UniLM and introduces additional weight. The memory updater depends on $W_a$, $W_b$, $W_c$ and $W_d$ to project memory and history values, as shown in Eq.(\ref{eq:mem-update}). Besides, the newly added special token \texttt{[E\_CLS]} and \texttt{[E\_SEP]} is supposed to learn appropriate embedding weight from scratch. It is challenging to expect the newly introduced weight can be learned properly if we directly fine-tune \ourapproach under the few shot scenario. 

Inspired by the pre-trained language models and the task adaptive pre-training~\cite{gururangan-etal-2020-dont}, we collect the unlabelled table side data to do a second phase task adaptive pre-training.

We first linearize the input table and add special token \texttt{[E\_CLS]} and \texttt{[E\_SEP]} to indicate the beginning and the end of the table slot value respectively. Then, around 20\% tokens are masked and the cross entropy loss is employed as the objective function. One corrupted example for further pre-training stage is ``\textit{\texttt{[CLS]} name is \texttt{[E\_CLS]} \texttt{[MASK]} kiprono \texttt{[MASK]} \texttt{[E\_SEP]}; birth\_date is \texttt{[E\_CLS]} 10 august \texttt{[MASK]} \texttt{[E\_SEP]}; \dots \texttt{[SEP]}}''.

During pre-training,
\ourapproach modifies the UniLM model architecture by designing a novel slot attention mask as well as slot memory mechanism which introduces additional weights. There are two goals for pre-training: 1) tune UniLM weights to incorporate slot attention mask , and 2) learn proper weights for slot memory block. We divide the pre-training stage into two phases: slot attention based pre-training and slot memory based pre-training.

We incrementally incorporate the slot attention and slot memory elements to the UniLM model along the two pre-training phases. First, the model structure of slot attention based pre-training is to add the slot attention mask to the last 6 layers of UniLM. We also learn the embedding of two special tokens \texttt{[E\_CLS]} and \texttt{[E\_SEP]} by adding them into the UniLM vocabulary. We load the UniLM checkpoint model weight as the initial weight for slot attention based pre-training. The second slot memory based pre-training phase adopts the full \ourapproach model, and is loaded with the checkpoint obtained after the slot attention mask based pre-training.

\subsection{Fine-Tuning and Inference}
In fine-tuning stage, \ourapproach first loads the model weight after the further pre-training stage which exploits valuable information from plenty of unlabelled task relevant data. The input for our proposed model is the concatenation of the linearized table and the reference sentence. The model is trained end to end in masked language model fashion. Around 70\% words in the reference are masked, and the cross entropy loss is used to minimize the discrepancy between the masked token and the groundtruth.

For inference, table side data is present while the reference sentence is missing. Our approach generates sentence auto-regressively. When making prediction on the $t$-th word, we need to inform the model previously generated table slots through table slot history representation $his^{t}$. 

\begin{table*}
\centering
\resizebox{0.98\textwidth}{!}{
\begin{tabular}{lllllll}
\bottomrule[1.5pt]
  &  & \textbf{BLEU-4} & \textbf{METEOR} & \textbf{ROUGE-L} & \textbf{PARENT(P/R/F)} & \textbf{PARENT-T(P/R/F)}\\
\hline
& \textbf{\textit{Humans}}  &   &      &   &     &   \\
\hline
1 & GPT2+copy \citep{chen-etal-2020-shot} & 41.7  & -  & -   & -  &  -  \\
2 & GPT2+copy (our replication) & 42.05  & 33.36  &   63.90 & 68.47/37.28/45.59   & 47.90/40.18/41.58   \\
3 & TableGPT2 \citep{gong-etal-2020-tablegpt}  &  45.6 & -  &  -  &  - &  -   \\
\hline
4 & GPT2 \citep{radford2019language} &  24.26  & 25.20 & 53.90  &  59.45/18.51/25.89   &  41.60/27.93/31.57  \\
5 & BART \citep{lewis-etal-2020-bart} & 48.31   &  37.24    & 68.24 &  74.04/41.46/50.79 &  \textbf{51.50}/41.98/44.20   \\
6 & UniLM \citep{dong2019unified} & 45.31  & 37.10  &   68.36   & 72.90/40.24/49.61  &   50.06/41.67/43.46  \\
7 & \ourapproach  &  \textbf{49.02} &  \textbf{37.97}   &  \textbf{69.37} &  \textbf{74.14}/\textbf{42.74}/\textbf{51.86}  &  51.20/\textbf{43.03}/\textbf{44.70} \\
\hline
 & \textbf{\textit{Books}}   &   &      &   &     &   \\
\hline
1 & GPT2+copy \citep{chen-etal-2020-shot} &  40.30 & -  &  -  & -  &   - \\
2 & GPT2+copy (our replication) & 40.39  &  34.48 & 67.59   & 69.68/35.10/44.87  &  51.34/35.34/40.45   \\
3 & TableGPT2 \citep{gong-etal-2020-tablegpt}  &  41.6 &  - & -   &  - & -    \\
\hline
4 & GPT2 \citep{radford2019language} & 19.12  &  24.99    &  54.83 &  55.22/17.72/24.94   & 40.41/28.21/32.14  \\
5 & BART \citep{lewis-etal-2020-bart} &  43.53 &  36.45    & 68.93  & 72.86/37.84/48.11  & \textbf{54.35}/\textbf{37.51}/\textbf{42.97}    \\
6 & UniLM \citep{dong2019unified} & 40.56  & 35.71 &   68.85    & 71.90/35.60/45.87  &  53.07/35.58/41.15    \\
7 & \ourapproach  & \textbf{43.88}  &  \textbf{36.98}  & \textbf{70.57}   &  \textbf{73.26}/\textbf{38.18}/\textbf{48.59} &   53.89/37.29/42.69  \\
\hline
 & \textbf{\textit{Songs}}   &   &      &   &     &   \\
\hline
1 & GPT2+copy \citep{chen-etal-2020-shot} &  42.20 &  - &  -  &  - &  -  \\
2 & GPT2+copy (our replication) &  42.41 &  33.43 & 65.18  & 66.34/35.72/44.75  & 42.05/33.99/36.27    \\
3 & TableGPT2 \citep{gong-etal-2020-tablegpt}  &  42.30 & -  & -   & -  &   -  \\
\hline
4 & GPT2 \citep{radford2019language} &  22.48 &  24.09    &  55.92 &  55.05/17.90/25.65   & 30.96/21.53/24.42  \\
5 & BART \citep{lewis-etal-2020-bart} & 43.88  &     34.69 & 67.22  &  \textbf{69.22}/36.31/46.00 &  \textbf{43.48}/34.55/37.26   \\
6 & UniLM \citep{dong2019unified} & 42.63  &  34.79    & \textbf{67.92}  &  68.19/34.74/44.55 & 41.32/32.64/35.24    \\
7 & \ourapproach  & \textbf{45.09}  &  \textbf{35.55}   & 67.38     & 67.60/\textbf{37.63}/\textbf{46.90}  &  42.78/\textbf{35.21}/\textbf{37.36} \\
\bottomrule[1.5pt]
\end{tabular}
}
\caption{\label{tab:humans-500}
Test results on three domains Humans/Books/Songs of Wiki dataset using 500 training data. ``P/R/F'' denotes the precision/recall/F score.
}
\end{table*}

\section{Experiment}
In this section, we explore the following experimental questions: (1) \textit{Can the proposed model generate fluent sentences?}; and (2) \textit{Is the generated sentence faithful to the fact given by input table?} We also perform ablation analysis to investigate the two main components of \ourapproach, namely the slot attention and slot memory mechanism.

\subsection{Dataset}

\paragraph{Task Adaptive Dataset for Pre-training} To pre-train \ourapproach, we collect additional unlabelled data from WikiBio \citep{lebret2016neural} and Wiki dataset. First, Wiki-Humans is a subset of WikiBio dataset which contains massive training examples collected from Wikipedia, a cleaned-up version of original WikiBio dataset by setting a vocabulary bound and removing those include out-of-vocabulary words that are not in the given table. Since pre-training only requires the table side data and focuses on reconstructing the corrupted text, we collect the rest of table side data (around 500K from WikiBio by removing all the train/valid/test data used in Wiki-Humans heuristically. Second, for songs and books domain, we collect around 26K and 17K filtered out table data from~\cite{chen-etal-2020-shot} respectively as the pre-training data.

\paragraph{Dataset for Fine-Tuning} Inspired by the experimental settings of few-shot natural language generation in~\cite{chen-etal-2020-shot}, we conduct experiments on three domains, i.e., humans, songs and books of Wiki dataset denoted as Wiki-Humans, Wiki-Songs and Wiki-Books. 
For each domain, we fine tune \ourapproach to inspect the model performance on various few shot settings by sampling different amount of training examples (e.g. 500, 200, 100, 50). The validation set for each domain includes 1000 instances, and test sets of humans, songs and books domain have 13587, 11879 and 5252 examples. We set the maximum length of the linearized table and the generated sentence as 300 and 64 respectively.

\subsection{Implementation Details}
The base model for \ourapproach is UniLM-base model with 12 Transformer layers, 768 hidden state dimensions, and 110M parameters in total. The implementation of \ourapproach is divided into two stages in total: 1) two-phase task adpative pre-training, and 2) fine-tuning on the target wiki dataset. We run the program on a single 1080Ti GPU with 12GB memory. Due to the memory constraint, the batch size on all stages is set as 4 and gradient is accumulated every 11 steps which results in a comparable 44 batch size. The learning rate is $5e\text{-}5$. The Adam~\cite{kingma2014adam} optimizer is used and the weight decay is set as 0.01.


For fine-tuning, we fine-tune the \ourapproach on target dataset by setting the maximum number of epoch as 50. For inference, we decode on the test set using the best checkpoints according to the validation set result. During inference, we use beam search with beam size 3 and length penalty 1.

\subsection{Baselines}
We compare the proposed model with strong pre-trained language models.
UniLM~\cite{dong2019unified} is a pre-trained language model for both natural language understanding and generation using three types of language modeling tasks.
BART~\cite{lewis-etal-2020-bart} introduces a denoising autoencoder for pre-training sequence-to-sequence models.
GPT-2~\cite{radford2019language} is a powerful unidirectional model pre-trained on millions of webpages in auto-regressive fashion. 
GPT2+copy~\cite{chen-etal-2020-shot} designed for few-shot table-to-text generation learns how to alternate between copying from table and generating functional words using GPT-2. TableGPT~\cite{gong-etal-2020-tablegpt} is a followup work of~\cite{chen-etal-2020-shot} while considers to minimize the contradicting part of the generated sentence give the table information. 

\subsection{Automatic Evaluation}

Following other generation tasks, we choose three automatic evaluation metrics BLEU-4~\cite{papineni2002bleu}, ROUGE-L~\cite{lin2004rouge} and METEOR~\cite{banerjee2005meteor} to evaluate the overlapping between the generated sentence and the reference sentence. Besides, to evaluate the faithfulness of generated sentence with the source table, we adopt PARENT~\cite{dhingra-etal-2019-parent} as our main metric. PARENT not only considers the matching between the generated sentence with the reference, but also takes how much table slot information is reflected in the generated sentence into account. In addition, to further evaluate the faithfulness of the generated text, PARENT-T~\cite{wang-etal-2020-towards} which only measures the matching between the generated text and the corresponding table is also included.

\begin{table*}
\centering
\resizebox{0.98\textwidth}{!}{
\begin{tabular}{lcccccccccccccc} 
\bottomrule[1.5pt]
Domain  & \multicolumn{4}{c}{Humans} &   & 
\multicolumn{4}{c}{Books}  &    & 
\multicolumn{4}{c}{Songs}\\ 
\cline{2-5}\cline{7-10}\cline{12-15} 
\# of training examples  & 50  & 100  & 200  &  500 &   
                        & 50  & 100  & 200  &  500 &  
                        & 50  & 100  & 200  &  500 \\
\hline
GPT2+copy (our replication) &  30.59 & 34.59  & 40.54  &  45.59 &  
                                &  42.67  & 42.79  & 43.44  &  44.87 &  
                                &  40.18  & 41.72  & 43.97 &  44.75 \\
\hline
GPT2 \citep{radford2019language}  &  0.17  &    12.90 & 19.02   & 25.89  & 
                                  & 0.71  & 20.82  & 24.18  &  24.94 & 
                                  &  0.85  & 17.08  & 24.72  &  25.65  \\
BART \citep{lewis-etal-2020-bart}  &  37.73  & 41.37 & 47.41 & 45.45  & 
                                  &  41.68  &  43.43 &  43.65 & 48.11 & 
                                  &   41.74 & 42.44  & 44.12  & 46.00   \\
UniLM \citep{dong2019unified}  &  35.80  & 41.83  & 46.08 &  49.61 & 
                              &  38.28  & 41.39  & 44.06  &  45.87 & 
                              &  40.17  & 41.95 & 42.45 &  44.55  \\
\hline
\ourapproach   & \textbf{43.55}  & \textbf{47.72}  & \textbf{50.13}  &  \textbf{51.86} & 
          &  \textbf{43.42}  & \textbf{46.03}  & \textbf{47.45}  &  \textbf{48.59} & 
          &  \textbf{42.03}  & \textbf{43.30}  & \textbf{45.93}  &  \textbf{46.90}  \\
\bottomrule[1.5pt]
\end{tabular}
}
\caption{
PARENT F score on three domains using 50/100/200/500 training examples.
}
\label{tab:parent-overall}
\end{table*}

\paragraph{Results}
We first compare \ourapproach with state-of-the-art models mentioned in section 4.3. Table~\ref{tab:humans-500} shows the performance of \ourapproach and baseline models on three domains of Wiki dataset using 500 training examples. For~\citep{chen-etal-2020-shot}, we copy the code that the author released on GitHub and replicate the result denoted as GPT2+copy (our replication). 
Regarding the conventional overlapping based metrics BLEU-4, METEOR, ROUGE-L, We can see that \ourapproach provides the best overall performance under various domains and evaluation metrics. \ourapproach outperforms the base model UniLM 3.71\%/3.32\%/2.46\% on BLEU-4 under Humans/Books/Songs domains, and \ourapproach gains 0.73\%/0.53\%/0.16\% more than the second best model BART on METEOR. \ourapproach outperforms the second best model BART 1.07\%/0.48\%/0.90\% on the F score of PARENT which is a strong indication that \ourapproach can achieve the strongest balance between the fluency and faithfulness. Regarding the overlapping between the generated sentence with table content, F scores of PARTENT-T metric shows that \ourapproach provides the most informative results on Humans and Songs domains while still very competitive with the best model BART on Books domain.

Besides, to verify the stability of \ourapproach when the amount of training data varies to 50, 100, 200 and 500, we show PARENT score for the proposed and other baseline models in Table~\ref{tab:parent-overall}. As shown in the table, over various domain and number of training example settings, \ourapproach outperforms other baseline models. Specifically, under the 200 training examples, \ourapproach outperforms the second strongest model BART by 2.72\% on Humans, UniLM by 3.39\% on Books, and BART by 1.81\% on Songs. The results demonstrate that leveraging the table slot attention as well as the memory mechanism provide a stable and competitive performance of faithful generation. On the other hand, on the Humans/Books/Songs domain with 50 training examples, \ourapproach gains 5.82\%/1.74\%/0.29\% improvements than the second best model BART respectively which shows that our model has powerful generative ability even only 50 examples are present. And human domains achieves the most gain since we collect most pre-training data for the task adaptive pre-training, thus it would be beneficial for the further work to collect more task adaptive pre-training data for Books and Songs domains to further boost the model performance.

\begin{table}
\centering
\resizebox{0.98\textwidth/2}{!}{
\begin{tabular}{lccccc} 
\bottomrule[1.5pt]
Domain  & \#sup & \#con  &  &  
overall \\
\cline{2-4} \cline{6-6}
Reference &  3.87  & 1.71   &   & 3.55   \\
\hline
GPT2+copy (our replication) &  3.99   & 1.75  &   & 3.39  \\
GPT2 \citep{radford2019language} &  3.73  & 1.69  &   & 3.61   \\
BART \citep{lewis-etal-2020-bart}  &  4.017  & \textbf{1.53}  &   & 3.24   \\
UniLM \citep{dong2019unified}  &  3.92  & 1.65  &   & 3.52  \\
\ourapproach &  \textbf{4.023}  & 1.75   &   & \textbf{3.22}   \\
\bottomrule[1.5pt]
\end{tabular}
}
\caption{\label{tab:human-evaluation}
Results of human evaluation.
}
\end{table}

\subsection{Analysis}
\begin{figure}[t]
\centering
\includegraphics[width=0.96\textwidth/2]{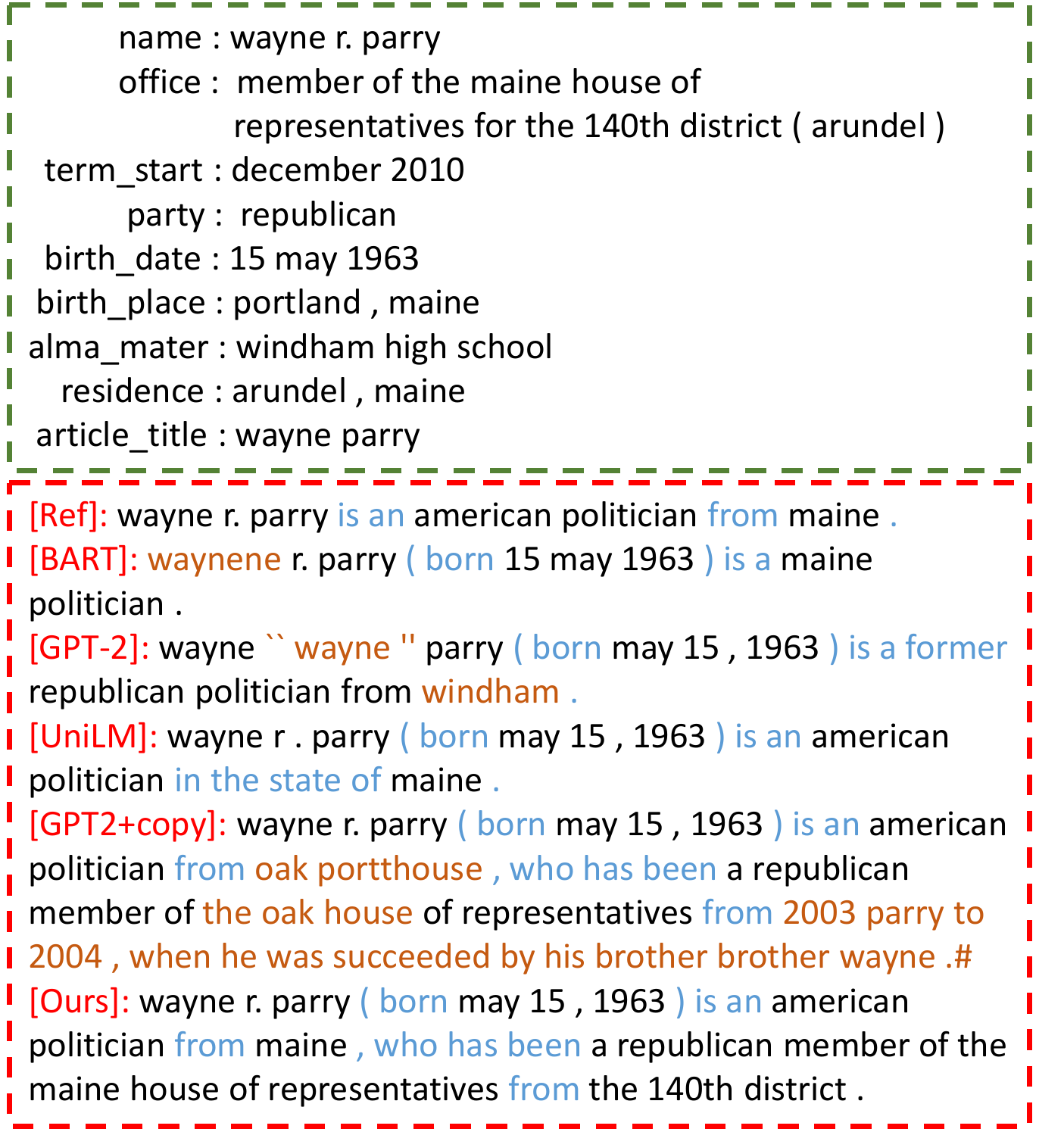}
\caption{A case study of a specific table input for qualitative analysis of table-to-text generation.}
\label{fig:case-study}
\centering
\end{figure}

We further analysis the faithfulness and the overall quality of the generated descriptions by conducting human evaluation. Then, we design ablation studies to investigate the importance of two building blocks of \ourapproach: span attention and memory mechanism.
In addition, we sample a specific input table and compare sentence generated by \ourapproach with the state-of-the-art models shown in Figure~\ref{fig:case-study}. 

\begin{table}[h]
\centering
\begin{tabular}{lccc} 
\bottomrule[1.5pt]
 & BART & AMG &   \\
 \hline
50 shots rating & 3.87  & 4.11 &  $p$ = 0.002 \\
500 shots rating & 4.46 & 4.55 &  $p$ = 0.24\\
\bottomrule[1.5pt]
\end{tabular}
\caption{\label{tab:human-evaluation-sig}
Statistical significance on human evaluation.
}
\end{table}

\paragraph{Human Evaluation} 
Following \citep{wang-etal-2020-towards,chen-etal-2020-shot}, we recruit three human annotators who pass the College English Test (CET-6) English test\footnote{A national English as a foreign language test in China.} to judge the quality of the generated sentence. We sample 100 test tables and collect corresponding outputs from \ourapproach, and baseline models. The sentences are randomly shuffled to reduce human variance. 
We provide instructions for human annotators to evaluate the sentence quality from two aspects: faithfulness and overall quality. First, for faithfulness, they are supposed to identify the number of entities mentioned in the sentence. Then, they need to compare the entities with ones from source table. Finally, they are supposed to report the number of fact supported and contradicted from the table respectively. Subsequently,  we compute the average number of supported and unsupported entities denoted by \#sup and \#con in Table~\ref{tab:human-evaluation}. 
The second study evaluates the overall quality of the generated sentence from their fluency, grammatical correctness, and the information consistency with the table. To compare the overall quality of various models, annotators rank the sentences generated using different models from 1 (best) to 6 (worst) by comparing the sentence. The ``overall'' column refers to the average ranking of the model.
Table~\ref{tab:human-evaluation} shows that \ourapproach generates better quality sentences compared with other models. Specifically, the outputs generated by \ourapproach contains the most information supported by the table and the overall quality is ranked the first place. Although it shows the number unsupported by the table is higher than other models, the overall quality still outperforms other models.

The overall ranking in Table~\ref{tab:human-evaluation} between BART and AMG is quite close, thus we ask 3 human evaluators to rate the generated sentences from 3 criteria, and then calculate the statistical significance of the overall rating between BART and AMG. We randomly sample 50 sentences for 50 and 100 training examples in few-shot cases respectively. Three annotators are instructed to re-evaluate the overall sentence quality by rating them from 1 (worst) to 5 (best) by considering the following 3 criteria: (1) \#sup,  (2) \#con (see Table 3), (3) naturalness and grammar correctness. The results are listed as follows.

As shown in Table~\ref{tab:human-evaluation-sig}, comparing BART with AMG, the p-value $p$ 0.002 of Wilcoxon signed-rank tests shows at $95\%$ confidence level, AMG is statistically significant with BART when training examples are as scarce as 50. While at $75\%$ confidence level, AMG is statistically significant with BART when training examples increase to 500.

\begin{table}[ht]
\centering
\resizebox{0.98\textwidth/2}{!}{
\begin{tabular}{l|cc|cc}
\cline{2-2}
\toprule[1.5pt]
\textbf{Model} & \textbf{BLEU} & \textbf{METEOR} & ~~ \textbf{PARENT} & ~~ \textbf{PARENT-T}\\ \hline
\ourapproach      &   \textbf{49.02}  &  37.97    & \textbf{51.86} &  \textbf{44.70}   \\ \hline 
\ourapproach w/o span &   47.28    &  37.10  & 50.24 &  43.36   \\ 
\ourapproach w/o mem   &  48.92   &  \textbf{38.14}     &  51.38    &  43.76  \\ 
\ourapproach w/o extra   &  46.78   &  36.99    &   49.83  &  44.00 \\

\bottomrule[1.5pt]
\end{tabular}
}
\caption{Ablation study of the proposed model.}
\label{tab:ablation}
\end{table}

\paragraph{Ablation Study} 
We also conduct ablation studies to understand each component of the proposed model, including slot attention and slot memory mechanism. Table~\ref{tab:ablation} provides the ablation results under different evaluation metrics.
It shows that \ourapproach can
still outperform all these two variants overall, certifying the effectiveness of each designed component in our model and we demonstrate that incorporating table slot attention and memory mechanism with the pre-trained model UniLM can boost the model performance.

\paragraph{Case Study} 

Figure~\ref{fig:case-study} provides a sample input table from test set along with various model outputs. The top box contains an input table while the bottom box includes model generations. In the bottom box, we leave the content supported by table as black, unsupported as light brown, and blue for the remaining words. We find that the output of pre-trained baseline models suffer from the following problems: (1) repetition, e.g., BART fails to generate person name ``\textit{wayne}'' correctly while repeats the last two letters as ``\textit{waynene}'', (2) hallucination, e.g., GPT-2 generates a middle name ``\textit{wayne}'' which is out of table, and GPT2+copy attempts to copy the ``\textit{office}'' slot but fail to copy the entire information by introducing unsupported information ``\textit{the oak house}'' and ``\textit{2003 ... brotherwayne.}''. By contrast, \ourapproach provides the highest table coverage while keeping the sentence fluent which demonstrates the table slot span attention and memory mechanism enables the model to copy from the table slot level correctly and enhance the generation faithfulness.

\section{Related Work}
\paragraph{Table-to-Text Generation} 

Recent years have witnessed much success on representing the semi-structured tabular data and generating text to describe the table.
From our investigation, most existing methods for table-to-text generation are based on the RNN-based encoder-decoder framework~\cite{lebret2016neural, liu2018table, wiseman2018learning, ma2019key, liu2019hierarchical}.
\citet{ma2019key} extend the table-to-text generation to low-resource scenario and put forward a Transformer-based model.
Of late, as the pre-training language model (e.g, BERT and GPT) has achieved significant successes in NLP, many works also propose to pre-train a model for table understanding.
\citet{yin20acl} pre-train a model for jointly understanding of tabular data around textual descriptions on large-scale paired data.
\citet{herzig-etal-2020-tapas} extend the architecture of BERT to encode tables as input, and propose a weakly supervised pre-training model for question answering over tables.
\citet{kale2020text} investigate the performance of pre-trained T5 \cite{2019t5} on multiple table-to-text tasks and provide a benchmark for the future research.
To keep the faithfulness of table on generation, one related work to ours is \cite{wang-etal-2020-towards}, which introduces a new table-text optimal-transport matching loss and a table-text embedding similarity loss based on the Transformer model to enforce the faithfulness during text generation.

\paragraph{Pre-Trained Language Model}

Our work is also related to model pre-training for NLP, which has brought dramatic improvements on natural language understanding~\citep{devlin2019bert,liu2019roberta,Clark2020ELECTRA,sun2019ernie} and generation~\citep{song2019mass,dong2019unified,liu2020commonsense,liu2019generative}. The widely used pre-trained models (PTMs) for table-to-text generation can be categorized into two classes: text-to-text PTMs~\citep{radford2018improving, devlin2019bert, dong2019unified, lewis-etal-2020-bart,joshi-etal-2020-spanbert} and structured data-to-text PTMs~\citep{chen-etal-2020-kgpt, herzig-etal-2020-tapas, xing-wan-2021-structure}. Recently, many pre-training models~\citep{liu2020kg,liu2020k,yao2019kg} start to incorporated the structured information from knowledge bases (KBs) or other structured semantic annotations into pre-training, which is also related to our work. 







\paragraph{Few-shot text generation}
Few-shot text generation learns with minimal data while maintaining decent generation capacity. Few-shot text generation can be used to augment the scarce training data to better assist the down-stream task, e.g., \cite{xia-etal-2020-composed,xia2020cg} for spoken language intent detection, \cite{brazinskas-etal-2020-shot} for opinion summary generation. In addition, to better utilize the available resources, \citet{chang-etal-2021-training} investigates the training instance selection on unlabelled data, and \citep{schick2020fewshot} adapts pattern-exploiting training strategy to fine-tune a PTM.

\section{Conclusion}
In this paper, we have proposed a novel approach 
\ourapproach for faithful table-to-text generation in few shots. 
We first attend over the multi-granularity of context using a novel span level and traditional token-by-token level attention strategy to exploit both the table structural and natural linguistic information.
Then, we design a memory unit to memorize the table slot allocation states dynamically. 
Extensive experiments on three domains of Wiki dataset verify the effectiveness of our proposed model on generating fluent and faithful descriptions from tables.

\section*{Acknowledgements}
We would like to thank all the anonymous reviewers for their helpful comments. This work is supported by NSF under grants III-1763325, III-1909323,  III-2106758, and SaTC-1930941. Yao Wan is partially supported by the Fundamental Research Funds for the Central Universities.

\bibliography{anthology}
\bibliographystyle{acl_natbib}




\end{document}